\documentclass[a4paper]{article}

\usepackage{amssymb,amsmath}
\usepackage{graphicx}

\usepackage{blindtext}
\usepackage[utf8]{inputenc}
\usepackage[english]{babel}
\usepackage{authblk}

\usepackage{url}
\urldef{\mailsa}\path|{anton.akusok,|
\urldef{\mailsb}\path|emil.eirola}@arcada.fi|

\usepackage{listings}
\graphicspath{{plots/}}

\newcommand{\x}{\ensuremath{\boldsymbol{x}}}
\newcommand{\y}{\ensuremath{\boldsymbol{y}}}

\newcommand{\X}{\ensuremath{\boldsymbol{X}}}

\newcommand{\W}{\ensuremath{\boldsymbol{W}}}
\newcommand{\bH}{\ensuremath{\boldsymbol{H}}}
\newcommand{\bbeta}{\ensuremath{\boldsymbol{\beta}}}

\begin{document}

\title{Comparison of Classification Methods for Very High-Dimensional Data in Sparse Random Projection Representation}

\author{Anton Akusok$^1$ \and Emil Eirola$^1$}
\date{%
    $^1$Department of Business Management and Analytics, Arcada UAS, Helsinki, Finland.\\%
}

\maketitle

\begin{abstract}
The big data trend has inspired feature-driven learning tasks, which cannot be handled by conventional machine learning models. Unstructured data produces very large binary matrices with millions of columns when converted to vector form. However, such data is often sparse, and hence can be manageable through the use of sparse random projections.

This work studies efficient non-iterative and iterative methods suitable for such data, evaluating the results on two representative machine learning tasks with millions of samples and features. An efficient Jaccard kernel is introduced as an alternative to the sparse random projection. Findings indicate that non-iterative methods can find larger, more accurate models than iterative methods in different application scenarios.
\end{abstract}

\section{Introduction}
\label{sec1}

Machine learning is a mature scientific field with lots of theoretical results, established algorithms and processes that address various supervised and unsupervised problems using the provided data. In theoretical research, such data is generated in a convenient way, or various methods are compared on standard benchmark problems -- where data samples are represented as dense real-valued vectors of fixed and relatively low length. Practical applications represented by such standard datasets can successfully be solved by one of a myriad of existing machine learning methods and their implementations.

However, the most impact of machine learning is currently in the big data field with the problems that are well explained in natural language (``Find malicious files'', ``Is that website safe to browse?'') but are hard to encode numerically. Data samples in these problems have distinct features coming from a huge unordered set of possible features. Same approach can cover a frequent case of missing feature values~\cite{EIROLA201432,SOVILJ2016220}. Another motivation for representing data by abstract binary features is a growing demand for security, as such features can be obfuscated (for instance by hashing) to allow secure and confidential data processing.

The unstructured components can be converted into vector form by defining indicator variables, each representing the presence/absence of some property (e.g., bag-of-words model~\cite{Harris1981}). Generally, the number of such indicators can be much larger than the number of samples, which is already large by itself. Fortunately, these variables tend to be sparse. In this paper, we study how standard machine learning solutions can be applied to such data in a practical way.

The research problem is formulated as a classification of sparse data with a large number of samples (hundreds of thousands) and huge dimensionality (millions of features). In this work, the authors omit feature selection methods because they are slow on such large scale, they can reduce the performance if a poor set of features is used, and, most importantly, features need to be re-selected if the feature set changes. Feature selection is replaced by Sparse Random Projection~\cite{srp2} (SRP) that provides a dense low-dimensional representation of a high-dimensional sparse data while almost preserving relative distances between data samples~\cite{srp1}. All the machine learning methods in the paper are compared on the same SRP representation of the data.

The paper also compares the performance of the proposed methods on SRP to find the suitable ones for big data applications. Large training and test sets are used, with a possibility to process the whole dataset at once. Iterative solutions are typically applied to large data processing, as their parameters can be updated by gradually feeding all available data. Such solutions often come at a higher computational cost and longer training time than methods where explicit solutions exist, as in linear models. Previous works on this topic considered neural networks and logistic regression~\cite{dahl2013large}. 
Also, there is application research~\cite{sayfullina2015efficient} without general comparison of classification methods. A wide comparison of iterative methods based on feature subset selection is given in the original paper for the publicly available URL Reputation benchmark~\cite{ma_identifying_2009}.

The remainder of this paper is structured as follows. The next section~\ref{sec:intro} introduces the sparse random projection, and the classification methods used in the comparison. The experimental section~\ref{sec:experiments} describes the comparison datasets and makes a comparison of experimental results. The final section~\ref{sec:conclusion} discusses the findings and their consequences for practical applications.

\section{Methodology}
\label{sec:intro}

\subsection{Sparse Random Projection for Dimensionality Reduction}
\label{sec:srp}

The goal of applying random projections is to efficiently reduce the size of the data while retaining all significant structures relevant to machine learning.
According to Johnson–Lindenstrauss' lemma, a small set of points can be projected from a high-dimensional to low-dimensional Euclidean space such that relative distances between points are well preserved~\cite{srp1}. As relative distances reflect the structure of the dataset (and carry the information related to neighborhood and class of particular data samples), standard machine learning methods perform well on data in its low-dimensional representation. The lemma requires an orthogonal projection, that is well approximated by random projection matrix at high dimensionality.

Johnson–Lindenstrauss lemma applies to the given case because the number of data samples is smaller than the original dimensionality (millions). However, computing such high-dimensional projection directly exceeds the memory capacity of contemporary computers. Nevertheless, a similar projection is obtained by using sparse random projection matrix. The degree of sparseness is tuned so that the result after the projection is a dense matrix with a low number of exact zeros.

Denote the random projection matrix by $\W$, and the original high-dimensional data by $\X$. The product $\W^T \X$ can be calculated efficiently for very large $\W$ and $\X$, as long as they are sparse. Specifically, the elements of $\W$ are not drawn from a continuous distribution, but instead distributed as follows:
\begin{equation}
w_{ij} =
\begin{cases}
-\sqrt{s/d} & \text{ with probability } 1 / 2s  \\
0 & \text{ with probability } 1 - 1 / s  \\
+\sqrt{s/d} & \text{ with probability } 1 / 2s
\end{cases}
\end{equation}
where $s=1 / \text{density}$ and $d$ is the target dimension~\cite{pingli}.

\subsection{Extreme Learning Machine}

{Extreme Learning Machine} (ELM)~\cite{huang_extreme_2006,huang_extreme_2012} is a single hidden layer feed-forward neural network where \emph{only} the output layer weights $\bbeta$ are optimized, and all the weights $w_{kj}$ between the input and hidden layer are assigned randomly. With $N$ input vectors $\x_i, \ i \in [1,N]$ collected in a matrix $\X$ and the targets collected in a vector $\y$, it can be written as
\begin{equation}
\bH \bbeta = \y \quad \text{where} \quad \bH = h(\W^T\X + \mathbf{1}^T\mathbf{b})
\label{eq:elm}
\end{equation}
Here $\W$ is a projection matrix with $L$ rows corresponding to $L$ hidden neurons, filled with normally distributed values, $\mathbf{b}$ is a bias vector filled with the same values, and $h(\cdot)$ is a non-linear activation function applied element-wise. This paper uses hyperbolic tangent function as $h(\cdot)$. Training this model is simple, as the optimal output weights $\bbeta$ are calculated directly by ordinary least squares. Tikhonov regularization~\cite{tikhonov_stability_1943} is often applied when solving the least square problem in Eq.~\eqref{eq:elm}. The value of the regularization parameter can be selected by minimizing the leave-one-out cross-validation error (efficiently calculated via the PRESS statistic~\cite{tropelm}). The model is easily adapted for sparse high-dimensional inputs by using sparse random matrix $\W$ as described in the previous section. ELM with this structure for the random weight matrix is very similar to the ternary ELM from~\cite{van2015binary}.

ELM can incorporate a linear part by attaching the original data features $\X$ to the hidden neurons output $\bH$. A random linear combination of the original data features can be used if attaching all the features is infeasible, as in the current case of very high-dimensional sparse data. These features let ELM learn any linear dependencies in data directly, without their non-linear approximation. Such method is similar to another random neural network method called Random Vector Functional Link network (RVFL~\cite{pao_94}), and is presented in this paper by the RVFL name.




\subsection{Radial Basis Function ELM}

An alternative way of computing the hidden layer output $\bH$ is by assigning a centroid vector $\mathbf{c}_j, \ j \in [1,L]$ to each hidden neuron, and obtain $\bH$ as a distance-based kernel between the training/test set and a fixed set of centroid vectors.
\begin{equation}
    \bH_{i,j} = e^{-\gamma_j d^2(\mathbf{x}_i, \mathbf{c}_j)}, \ i \in [1,N], \ j \in [1,L],
\end{equation}
\noindent where $\gamma_j$ is kernel width.

Such architecture is widely known as Radial Basis Function (RBF) network~\cite{broomhead1988multivariable,lowe1989adaptive}, except that ELM-RBF uses fixed centroids and fixed random kernel widths $\gamma_j$. Centroid vectors $\mathbf{c}_j$ are chosen from random training set samples to better follow the input data distribution. Distance function for dense data is Euclidean distance.

\subsection{Jaccard distance for sparse binary data}

Distance computations are a major drawback in any RBF network with Euclidean distances as they are slow and impossible to approximate for high-dimensional data~\cite{beyer_when_1999}. Jaccard distances can be used for binary data~\cite{czarnecki_weighted_2015}. However, a naive approach for Jaccard distances is infeasible for datasets with millions of features.

An alternative computation of Jaccard distance matrix directly from sparse data is considered in the paper and proved to be fast enough for practical purposes. Recall the Jaccard distance formulation for sets $a$ and $b$:
\begin{equation}
    J(a,b) = 1 - \frac{|a \cap b|}{|a \cup b|}
\end{equation}

Each column in sparse binary matrices $\mathbf{A}$ and $\mathbf{B}$ can be considered as a set of non-zero values, so $\mathbf{A} = [a_1, a_2, \ldots a_m]$ and  $\mathbf{B} = [b_1, b_2, \ldots b_n]$. Their union and intersection can be efficiently computed with matrix product and reductions:
\begin{eqnarray}
    |a_i \cap b_j| &=& (\mathbf{A}^T\mathbf{B})_{ij}, \ i \in [1,n], \ j \in [1,m]\\
    |a_i \cup b_j| &=& |a_i| + |b_j| - |a_i \cap b_j| = \left( \mathbf{1}^T\sum_k{A_{ik}} + \sum_K{B_{jk}}^T\mathbf{1} - \mathbf{A}^T\mathbf{B} \right)_{i,j}
\end{eqnarray}

The sparse matrix multiplication is the slowest part, so this work utilizes its parallel version. Note that the runtime of a sparse matrix product $\mathbf{A}^T\mathbf{B}$ scales sub-linearly in the number of output elements $n\cdot m$, so the approach is inefficient for distance calculation between separate pairs of samples $(a_i, b_j)$ not joined in large matrices $\mathbf{A}$, $\mathbf{B}$.

\section{Experiments}
\label{sec:experiments}

\subsection{Datasets}

The performance of the various methods is compared on two separate, large datasets with related classification tasks.
The first dataset concerns Android application packages, with the task of malware detection. Features are extracted using static analysis techniques, and the current data consists of 6,195,080 binary variables. There are 120,000 samples in total, of which 60,000 are malware, and this is split into a training set of 100,000 samples and a fixed test set of 20,000 samples.

The data is very sparse -- the density of nonzero elements is around 0.0017\%. Even though the data is balanced between classes, the relative costs of false positives and false negatives are very different. As such, the overall classification accuracy is not the most useful metric, and the area under the ROC curve (AUC) is often preferred to compare models. More information about the data can be found in~\cite{sayfullina2015efficient,sayfullina2016android,palumbo_pragmatic}.

Second, the Web URL Reputation dataset~\cite{ma2009identifying} contains a set of 2,400,000 websites that can be malicious or benign. The dataset is split into 120 daily intervals when the data was collected; the last day is used as the test set. The task is to classify them using the given 3,200,000 sparse binary features, as well as 65 dense real-valued features. This dataset has 0.0035\% nonzero elements, however, a small number of features are real-valued and dense. For this dataset, the classification accuracy is reported in comparison with the previous works~\cite{zhang2015tree,zhou2016one}.

\subsection{Additional Methods}

Additional methods include Kernel Ridge Regression (KRR), $k$-Nearest Neighbors (kNN), Support Vector Machine for binary classification (SVC), Logistic regression and Random Forest. Of these methods, only SVC and logistic regression have iterative solutions.

Kernel Ridge Regression~\cite{saunders1998ridge,murphy_machine_2012} combines Ridge regression, a linear least squares solution with L2-penalty on weights norm, with the kernel trick. Different kernels may be used, like the Jaccard distance kernel for sparse data proposed above.

$k$-Nearest Neighbors (kNN) method is a simple classifier that looks at $k$ closest training samples to a given test sample and runs the majority vote between them to predict a class. The value of $k$ is usually odd to avoid ties. It can use different distance functions, and even a pre-computed distance matrix (with the Jaccard distance explained in the Methodology section).

Support Vector Machine~\cite{cortes1995support} constructs a hyperplane in a kernel space, that separates the classes. It's a native binary classifier with an excellent performance, that can be extended to regression as well. There is a significant drawback of the computational complexity that is between quadratic and cubic in the number of samples~\cite{svm_comp}.

Logistic regression~\cite{christopher2006pattern} is a binary classifier that utilizes the logistic function in its optimization problem. The logistic function is non-linear and prevents Logistic Regression from having a direct one-step solution like linear least squares in ELM or RVFL. Its weights are optimized iteratively~\cite{schmidt2013minimizing}, with an L2 penalty helping to speed up the convergence and improve the prediction performance.

Random Forest~\cite{breiman2001random,zhang2014towards} is an ensemble method consisting of many random tree classifiers (or regressors). Each tree is trained on bootstrapped data samples and a small subset of features, producing a classifier with large variance but virtually no bias. Then Random Forest averages the predictions of all trees, reducing the variance of their estimations. It can be considered a non-iterative method because parameters of each leaf of a tree are computed in a closed form and never updated afterward. The method can work directly with sparse high-dimensional data. It obtains good results already with 10 trees. Increasing this number yields little improvement at a significant price in the training time. Drawbacks of Random Forest are its inability to generalize beyond the range of feature values observed in the training set, and slower training speed with a large number of trees.

All methods are implemented in Python using scikit-learn routines~\cite{scikit-learn}. The experiments are run on the same workstation with a 6-core Intel processor and 64GB RAM.

\subsection{Results on Large Data}

The effect of various Sparse Random Projection dimensionality on classification performance for large datasets is examined here. A range of 4--10,000 features is evaluated, sampled uniformly on a logarithmic scale. Only ``fast'' methods with sub-quadratic runtime scaling in the number of training samples are considered.

\begin{figure}[ht]
    \centering
    \includegraphics[width=0.495\textwidth]{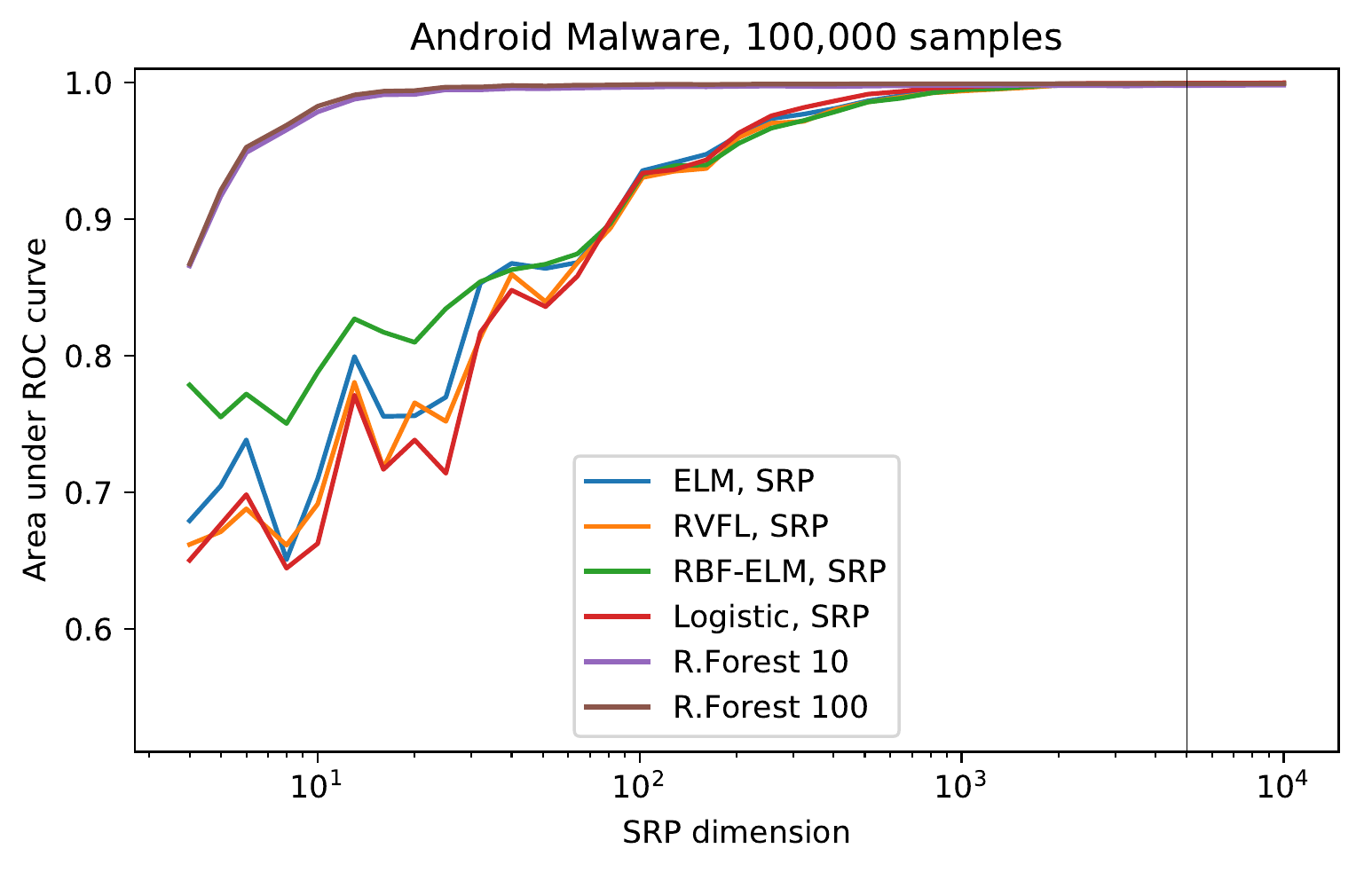}
    \includegraphics[width=0.495\textwidth]{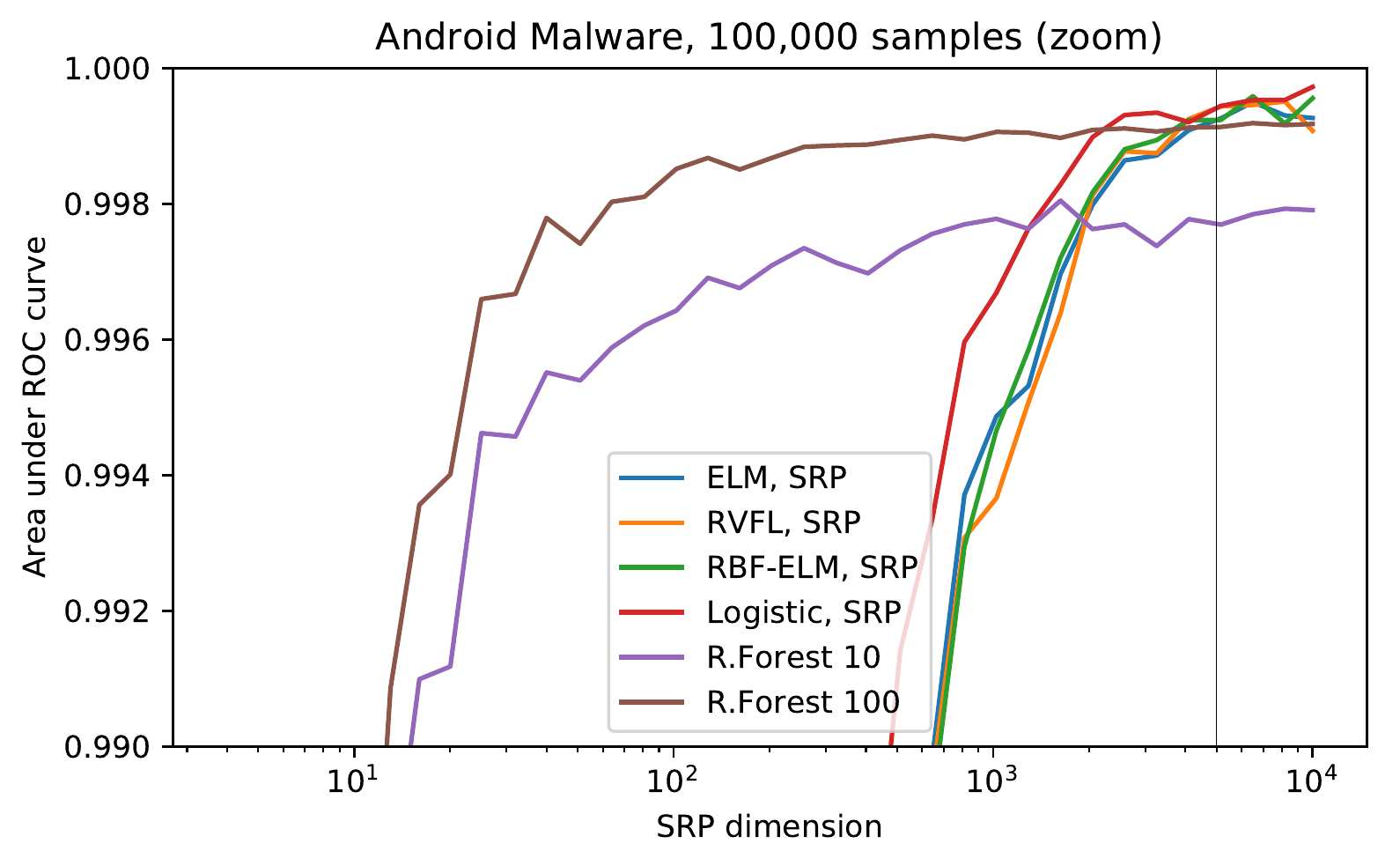}
    \caption{The effect of varying number of Sparse Random Projection features on the performance for Android Malware dataset. Vertical line corresponds to 5000 features.}
    \label{fig:all1}
\end{figure}

Performance evaluation for Android Malware dataset is presented on Figure~\ref{fig:all1}. Random Forest methods are the best performers with little SRP features, but other methods catch up after 2,000 features in SRP and outperform Random Forest with even higher SRP dimensionality. The number of trees in Random Forest has a small positive effect of only +0.2\% AUC for 10 times more trees.

\begin{figure}[ht]
    \centering
    \includegraphics[width=0.495\textwidth]{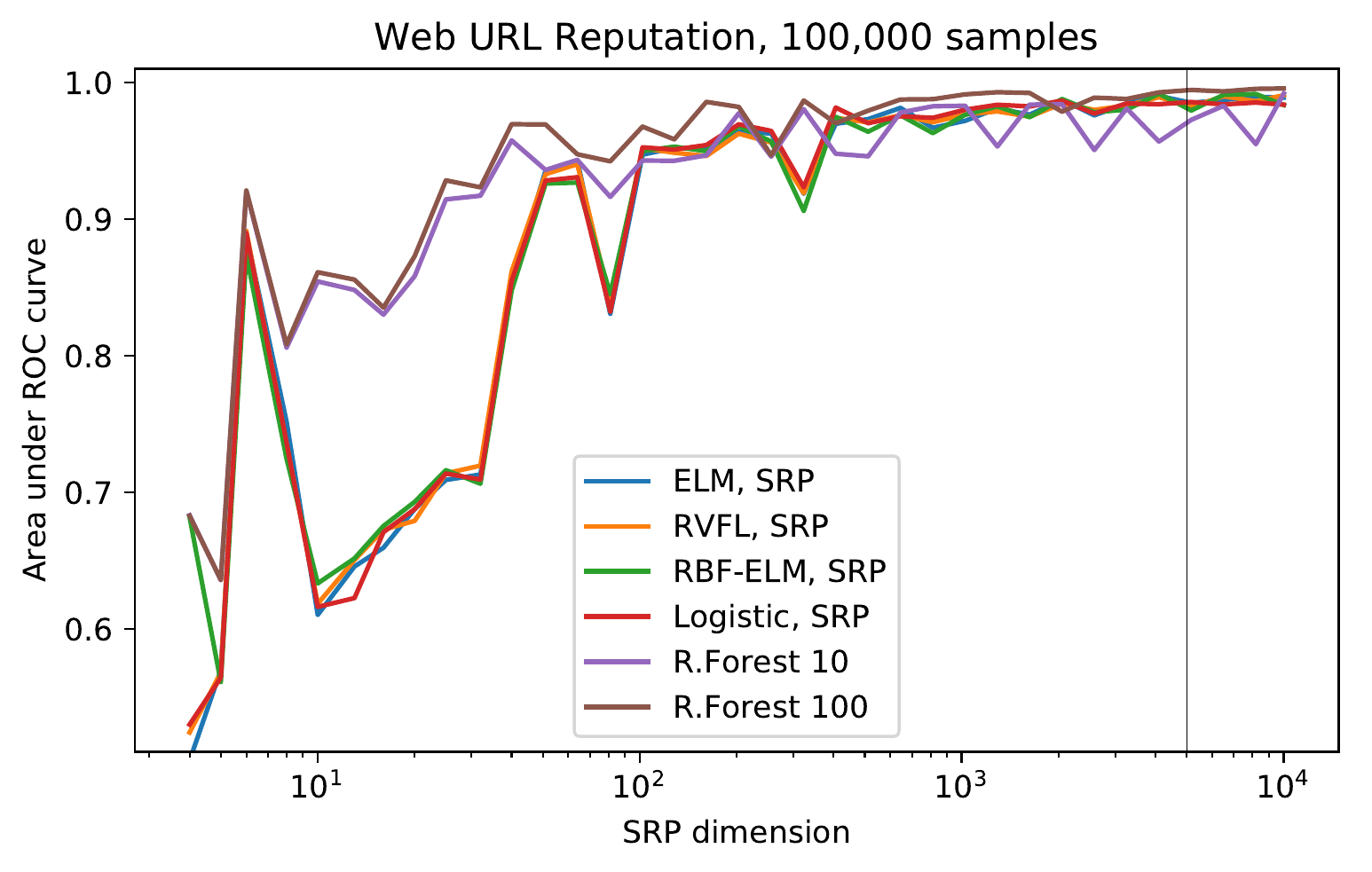}
    \includegraphics[width=0.495\textwidth]{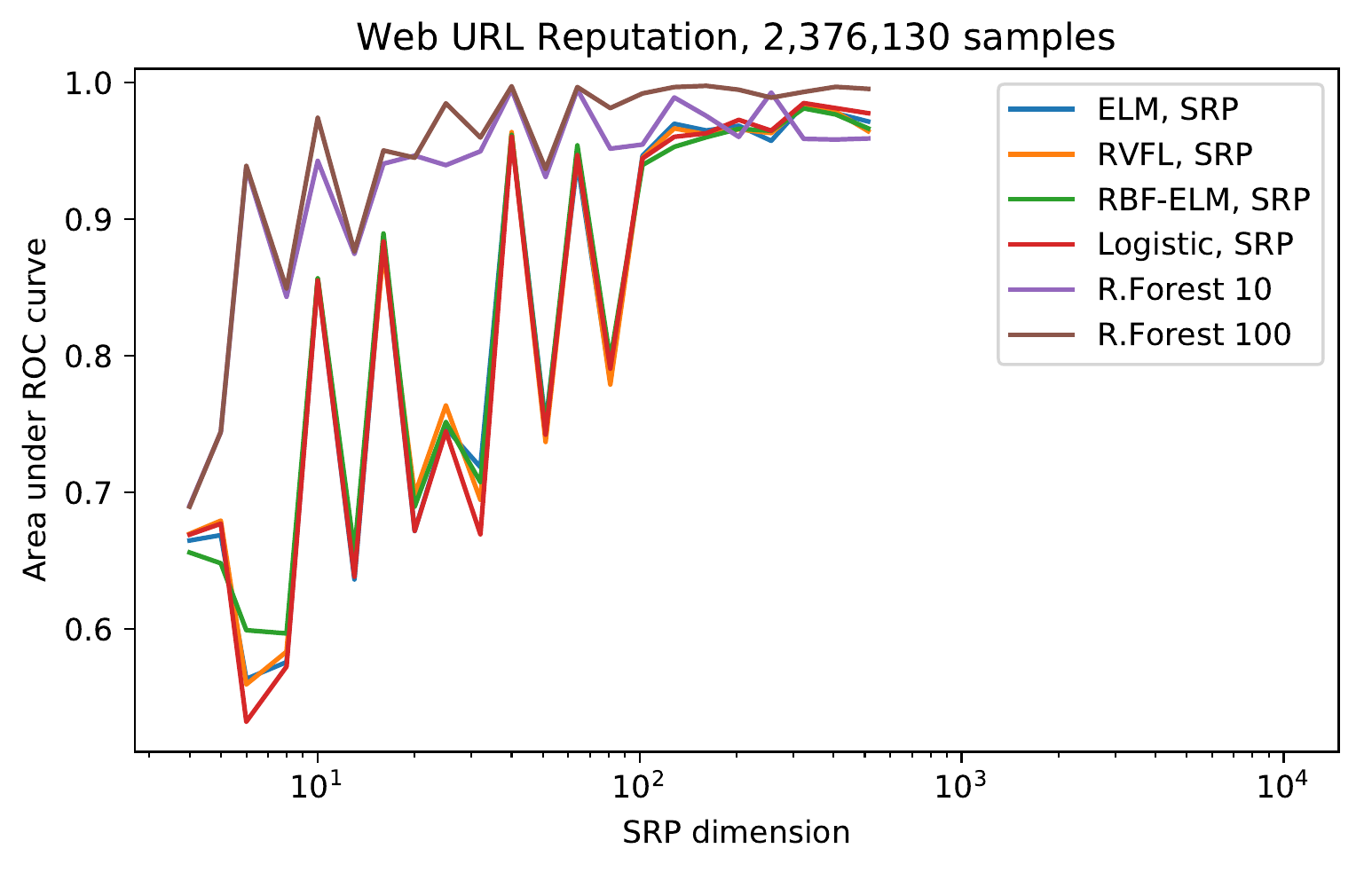}
    \caption{The effect of varying number of Sparse Random Projection features on the performance for Web URL Reputation dataset. Full dataset is tested with a maximum 650 features due to the memory constraints. Vertical line corresponds to 5,000 features.}
    \label{fig:all2}
\end{figure}

The same evaluation for Web URL Reputation dataset is shown in Figure~\ref{fig:all2}. Random Forest is again better with fewer SRP features and performs similarly with more features. More trees in Random Forest reduce is performance fluctuations. Interesting that all the methods except Random Forest perform very similarly despite their different formulation; a fact that is probably connected to the nature of Sparse Random Projection.



\subsection{Sparse Random Projection Benchmark}

A larger variety of classification methods are compared on a reduced training set of only 1,000 samples, randomly selected from the training set. All experiments use the same test set of 20,000 samples. Sparse Random Projection includes 5,000 features that provide highest performance with reasonable runtime in the previous experiments (vertical line on Figures~\ref{fig:all1}, \ref{fig:all2}). The experimental results are summarized in Table~\ref{tab:1}.

A total of 101 fixed training sets are generated -- one for tuning hyper-parameters, and the rest for 100 runs of all the methods. All methods use tuned L2-regularization with the regularization parameter selected on a logarithmic scale of $[2^{-20}, 2^{-19}, \ldots, 2^{20}]$, except for kNN that has a validated $k=1$ and Random Forest with 100 trees for a reasonable runtime.

Comparison results on Web URL Reputation dataset achieved 94.4\% accuracy in rule-based learning~\cite{zhang2015tree}, and 97.5\% accuracy in SVM-based approach~\cite{ma2009identifying}. The latter is comparable to the proposed results, but the exact comparison depends on a particular point of the ROC curve.


\begin{table}[ht]
\caption{Mean area under ROC curve in \% (with the standard deviation in parentheses) and runtime in seconds for all methods on the two benchmark datasets, using 1,000 training samples and summarized over 100 runs.
Bold font denotes the best result for each dataset, and any other not statistically significantly different values (paired t-test at the significance level 0.05).}
\label{tab:1}
\centering
\bigskip

\begin{tabular}{l|c|c|c|c}
Method & \multicolumn{2}{|c}{Android Malware} &  \multicolumn{2}{|c}{Web URL Reputation} \\
\hline
 & AUC (std.), \% & time, s & AUC (std.), \% & time, s \\
 \hline

ELM, SRP &  99.41 (0.08)  & \phantom03.1  & \bf 99.29 (0.16)  & \phantom02.9 \\
RVFL, SRP &  99.34 (0.08)  & \phantom02.2  &  98.13 (0.74)  & \phantom02.1 \\
RBF-ELM, SRP &  99.12 (0.11)  & \phantom02.1  &  97.53 (1.81)  & \phantom02.1 \\
KRR, SRP &  92.61 (0.24)  & \phantom01.3  &  99.15 (0.22)  & \phantom01.3 \\
kNN, SRP &  86.55 (1.00)  & 13.4  &  83.35 (2.25)  & 13.5 \\
SVC, SRP &  99.26 (0.10)  & 44.2  &  99.17 (0.21)  & 34.6 \\
Logistic Regression, SRP &  99.21 (0.11)  & \phantom00.6  &  99.17 (0.23)  & \phantom00.1 \\
Random Forest &  98.54 (0.42)  & 30.6  &  95.64 (1.65)  &  \phantom04.9 \\
RBF-ELM, Jaccard &  86.34 (5.15)  & 18.9  &  79.67 (3.62)  & \phantom01.7 \\
KRR, Jaccard & \bf 99.48 (0.06)  & 18.3  & \bf 99.31 (0.06)  & \phantom01.6 \\
kNN, Jaccard &  91.38 (0.50)  & 18.0  &  84.62 (0.95)  & \phantom01.3 \\
\hline
\end{tabular}
\end{table}

\section{Conclusion}
\label{sec:conclusion}

This study provides useful insights on the nature of a very high-dimensional sparse data and the utility of Sparse Random Projection for its processing.

The original high-dimensional sparse data representation is best combined with Random Forest if the data is abundant. Random Forest efficiently learns to discriminate between classes on a huge dataset and has the fastest training speed if run in parallel with a small number of trees. However, it underperforms on smaller datasets.

The original sparse data can be used in kernel matrix calculation for Kernel Ridge regression, that excels in smaller datasets. However, the kernel computation runtime is a significant drawback, and the KRR itself cannot scale to huge datasets.

Sparse Random Projection efficiently represents a high-dimensional sparse data given a sufficient number of features (at least 1,000 in the tested datasets). In that case, it provides good results with very different methods: based on neural networks, logistic regression, kernel methods like KRR and SVM. An interesting fact is that the choice of a particular method is not significant. Of the aforementioned methods, ELM and RVFL are the most versatile. They provide best results in a short runtime, for any training set size.

\section*{Acknowledgements}

This work was supported by Tekes -- the Finnish Funding Agency for Innovation -- as part of the ``Cloud-assisted Security Services'' (CloSer) project.

\bibliographystyle{plain}
\bibliography{sparse_elm}

\end{document}